\documentclass[10pt,twocolumn,letterpaper]{article}

\usepackage{wacv}
\usepackage{times}
\usepackage{epsfig}
\usepackage{graphicx}
\usepackage{amsmath}
\usepackage{amssymb}

\usepackage{overpic}

\usepackage{color}
\usepackage[normalem]{ulem}
\usepackage{float}
\usepackage{tabularx}
\usepackage{booktabs}
\usepackage{dsfont}
\usepackage{subfigure}
\usepackage{multirow}

\usepackage[pagebackref=true,breaklinks=true,letterpaper=true,colorlinks,bookmarks=false]{hyperref}

\wacvfinalcopy 

\ifwacvfinal\fi
\newcommand{\comment}[1]{}

\newcommand{\bI}{\mathbf{I}}

\newcommand{\bm}{\mathbf{m}}

\newcommand{\bT}{\mathbf{T}}

\newcommand{\bh}{\mathbf{h}}
\newcommand{\bhh}{\hat{\mathbf{h}}}

\newcommand{\IR}{\mathds{R}}

\newcommand{\template}{\text{template}}
\newcommand{\templates}{\text{templates}}
\newcommand{\view}{\text{view}}
\newcommand{\errnet}[0]{registration error network\xspace}
\newcommand{\regnet}[0]{initial registration network\xspace}

\newcommand{\bPhi}{{\boldsymbol{\Phi}}}
\newcommand{\bPsi}{{\boldsymbol{\Psi}}}

\newcommand{\fig}[1]{Fig.~\ref{fig:#1}}

\newcommand{\tbl}[1]{Table~\ref{tbl:#1}}

\newcommand{\refsec}[1]{Section~\ref{sec:#1}}

\newcommand{\cW}{\mathcal{W}}

\newcommand{\err}{\text{Err}}
\newcommand{\IoUwhole}{\text{IoU}_\text{whole}}
\newcommand{\IoUpart}{\text{IoU}_\text{part}}

\newcommand{\loss}{\mathcal{L}}

\renewcommand{\paragraph}[1]{\vspace{0.2em}\noindent\textbf{#1}}

\makeatletter
\newcommand\footnoteref[1]{\protected@xdef\@thefnmark{\ref{#1}}\@footnotemark}
\makeatother 

\begin{document}

\title{Optimizing Through Learned Errors for Accurate Sports Field Registration}

\author{Wei Jiang\textsuperscript{1} \qquad Juan Camilo Gamboa Higuera\textsuperscript{2,3} \qquad Baptiste Angles\textsuperscript{1} \\ Weiwei Sun\textsuperscript{1} \qquad Mehrsan Javan\textsuperscript{3} \qquad Kwang Moo Yi\textsuperscript{1}
\vspace{0.5em}
\\
{\small \textsuperscript{1}Visual Computing Group, University of Victoria\qquad
\textsuperscript{2}McGill University\qquad
\textsuperscript{3}SPORTLOGiQ Inc.}\\
{\tt\small \{jiangwei, bangles, weiweisun, kyi\}@uvic.ca, \{gamboa\}@cim.mcgill.ca, \{mehrsan\}@sportlogiq.com}
}

\maketitle
\ifwacvfinal\thispagestyle{empty}\fi

\begin{abstract}
We propose an optimization-based framework to register sports field  \templates{} onto broadcast videos.
For accurate registration we go beyond the prevalent feed-forward paradigm.
Instead, we propose to train a deep network that regresses the registration error, and then register images by finding the registration parameters that minimize the regressed error.
We demonstrate the effectiveness of our method by applying it to real-world sports broadcast videos, outperforming the state of the art.
We further apply our method on a synthetic toy example and demonstrate that our method brings significant gains even when the problem is simplified and unlimited training data is available.
\footnote{
Code is available at \url{https://github.com/vcg-uvic/sportsfield_release}.
}
\end{abstract}

\section{Introduction}
\label{sec:intro}

Estimating the relationship between a \template{} and an observed image with deep learning~\cite{Detone16,Rad17,Kehl17,Tekin18a} has received much attention recently, due to the success of deep learning in many other areas in computer vision~\cite{He15,He17,Ronneberger15}.
Registration of a sports field \template{} onto a camera view is not an exception~\cite{Chen18e, Homayounfar17d,Sharma18d}, where deep learning has shown promising results compared to traditional baselines.
Despite the recent advancements, there is room for further improvement, especially for augmented reality and sport analytics.

For mixed and augmented reality, even the slightest inaccuracies in estimates can break immersion~\cite{Marchand16}.
For sports analytics, good alignment is crucial for detecting the important events -- \eg offsides in soccer.

Existing methods have also acknowledged this limitation, and have sought to improve accuracy. 
For example, some rely on a hierarchical strategy~\cite{Rad17,Kehl17}.
In these methods, the refinement network is used on top of a rough pose estimator, where both are feed-forward networks. 
However, as we will demonstrate through our experiments, there is an alternative way to enhance performance.

In order to achieve more accurate registration, we take a different route to the commonly used feed-forward paradigm.
Inspired by classic optimization-based approaches for image registration~\cite{Puwein11,Lucas81, Moreno09a, Oberkampf96}, we propose optimizing to reduce the estimated registration error.
Opposed to traditional methods, we rely on a deep network for estimating the error.

\begin{figure*}
\includegraphics[width=1.0\linewidth]{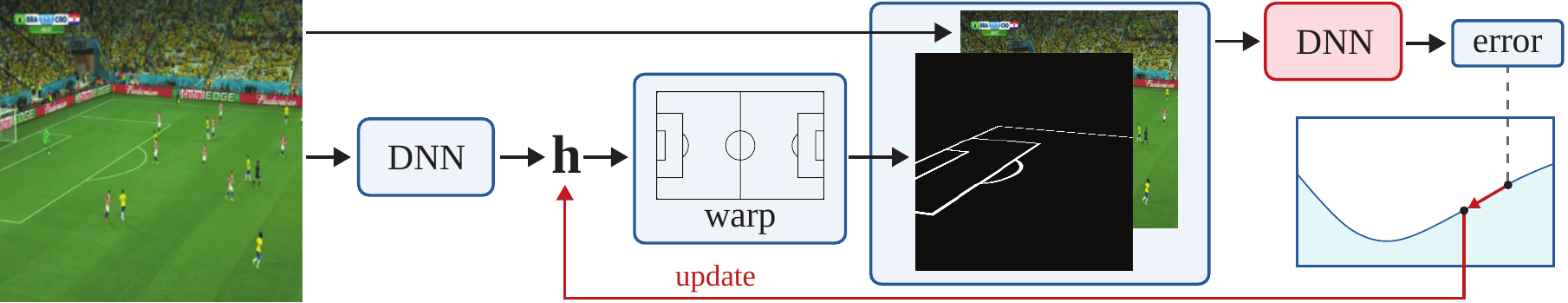}
\caption{
Illustration of our framework.
We train two deep networks each dedicated for a different purpose.
We first obtain an initial pose estimate from a feed-forward network that regresses directly to the homography parameterization $\bh$ -- DNN on the left in blue.
We then warp our sports field \template{} according to this initial estimate and concatenate it with the input image.
We feed this concatenated image to a second deep neural network -- DNN on the right in red -- that estimates the error of the current warping.
We then differentiate through this network to obtain the direction in which the estimated error is minimized, and optimize our estimated homography accordingly -- red arrow.
This optimization process is repeated multiple times until convergence.
All figures in this paper are best viewed in color.
}
\label{fig:framework}
\end{figure*}

Specifically, as illustrated in \fig{framework}, we propose a two-stage deep learning pipeline, similar to existing methods~\cite{Rad17,Kehl17}, but with a twist on the refinement network.
The first-stage network, which we refer to as the \emph{\regnet}, provides a rough estimate of the registration, parameterized by a homography transform.
For the second-stage network, instead of a feed-forward refinement network, we train a deep neural network that regresses the error of our estimates -- \emph{\errnet}.
We then use the \regnet to provide an initial estimate, and optimize the initial estimate using the gradients provided by differentiating through the \errnet.
This allows much more accurate estimates compared to the single stage feed-forward inference.

In addition, we propose \emph{not} to train the two networks together. 
While end-to-end and joint training is often preferred~\cite{Ren15,Ono18}, we find that it is beneficial to train the two networks separately -- decoupled training.
We attribute this to two observations:
the two networks -- \regnet and \errnet ~-- aim to regress different things -- pose and error; it is useful for the \errnet to be trained independently of the \regnet so that it does not overfit to the mistakes the \regnet makes while training.

We demonstrate empirically that our framework consistently outperforms feed-forward pipelines.
We apply our method to sports field registration with broadcast videos.
We show that not only our method outperforms feed-forward networks in a typical registration setup, it is also able to outperform the state of the art even when training data is scarce -- a trait that is desirable with deep networks.
We further show that our method is not limited to sport fields registration.
We create a simple synthetic toy dataset of estimating equations of a line in the image, and show that even in this simple case when unlimited train data is available, our method brings significant advantage.

To the best of our knowledge, our method is the first method that learns to regress registration errors for optimization-based image registration.
The idea of training a deep network to regress the error and perform optimization-based inference has recently been investigated for image segmentation, multi-label classification, and object detection~\cite{Gygli17, Jiang18b}. 
While the general idea exists, it is non-trivial to formulate a working framework for each task.
Our work is the first successful attempt for sport field registration, thanks to our two-stage pipeline and the way we train the two networks.

To summarize, our contributions are: 
\begin{itemize}
    \item we propose a novel two-stage image registration framework where we iteratively optimize our estimate by differentiating through a learned error surface;
    \item we propose a decoupled training strategy to train the \regnet and the \errnet;
    \item our method achieves the state-of-the-art performance, even when the training dataset size is as small as 209 images;
    \item we demonstrate the potential of our method through a generic toy example.
    
\end{itemize}

\section{Related Work}
\label{sec:related}

\paragraph{Sports field registration.}
Early attempts on registering sports field to broadcast videos~\cite{Gupta11, Ghanem12, Puwein11} typically rely on a set of pre-calibrated reference images. 
These calibrated references are used to estimate a relative pose to the image of interest.
To retrieve the relative pose, these methods either assume that images are of correspond to consecutive frames in a video~\cite{Gupta11,Ghanem12}, or use local features, such as SIFT~\cite{Lowe04} and MSER~\cite{Matas04}, to find correspondences~\cite{Puwein11}.
These methods, however, require that the set of calibrated images contains images with similar appearance to the current image of interest, as traditional local features are weak against long-term temporal changes~\cite{Verdie15}.
While learned alternatives exist~\cite{Yi16b,Ono18,DeTone17b}, their performances in the context of pose estimation and registration remains questionable~\cite{Schonberger17}.

To overcome these limitations, more recent methods~\cite{Homayounfar17d, Sharma18d, Chen18e} focus on converting broadcast videos into images that only contain information about sports fields, \eg known marker lines, then perform registration.
Homayounfar \etal~
\cite{Homayounfar17d} perform semantic segmentation on broadcast images with a deep network, then optimize for the pose using branch and bound~\cite{Lampert09} with a Markov Random Field (MRF) formulated with geometric priors.
While robust to various scenic changes, their accuracy is still limited.
Sharma \etal~
\cite{Sharma18d} simplify the formulation by focusing on the edges and lines of sports fields, rather than the complex semantic segmentation setup.
They use a database of edge images generated with known homographies to extract the pose, which is then temporally smoothed.
Chen and Little~
\cite{Chen18e} further employ an image translation network~\cite{Isola17} in a hierarchical setup where the play-field is first segmented out, followed by sports field line extraction.
They also employ a database to extract the pose, which is further optimized through Lucas-Kanade optimization~\cite{Lucas81} on distance transformed version of the edge images.
The bottleneck of these two methods is the necessity of a database, which hinders their scalability.

\paragraph{Homography estimation between images.}
Traditional methods for homography estimation include sparse feature-based approaches~\cite{Yan14} and dense direct approaches \cite{Lucas81}.
Regardless of sparse or dense, traditional approaches are mainly limited by either the quality of the local features~\cite{Wu07a}, or by the robustness of the objective function used for optimization~\cite{Baker04b}.

Deep learning based approaches have also been proposed for homography estimation.
In \cite{Detone16}, the authors propose to train a network that directly regresses to the homography between two images through self supervision. Interestingly, the output of the regression network is discretized, allowing the method to be formulated as classification.
Nguyen \etal~\cite{Nguyen18d} train a deep network in an unsupervised setup to learn to estimate the relative homography.
The main focus of these methods, however, is on improving the inference speed, without significant improvements on accuracy when compared
to traditional baselines.

\paragraph{Feed-forward 6 Degree-of-Freedom (DoF) pose estimators.}
Pose estimators are also higly related to image registration.
Deep networks have also been proposed to directly regress the 6 DoF pose of cameras~\cite{Xiang17, Mahendran17, Kendall15a}.
Despite being efficient to compute, these methods highly depend on their parameterization of the pose -- naive parameterizations can lead to bad performance,
and are known to have limited accuracy~\cite{Tekin18}.
To overcome this limitation, recent works focus on regressing the 2D projection of 3D control points~\cite{Tekin18a, Rad17, Kehl17}. 
Compared with directly predicting the pose, control points based pose show improved performance due to the robust estimation of parameters.
Our initial registration network follows the same idea as these methods to obtain our initial estimate.

\paragraph{Optimizing with learned neural networks.}
Incorporating optimization into deep pipelines is a current topic of interest. BA-Net \cite{Tang18} learns to perform Levenberg-Marquardt optimization within the network to solve dense bundle adjustment.
LS-Net \cite{Clark18} learns to predict the directions to improve a given camera pose estimate.
Han \etal~\cite{Han18} also learn to estimate the Jacobian matrix from an image pair to update the 6 DoF camera pose. In contrast to these methods, which propose learning a function to update a camera pose estimate, we propose to learn an error function that predicts how well two images are aligned. Using the error function we can obtain the update direction via differentiation.
The most similar work to ours is the deep value networks~\cite{Gygli17},
where they train a network to estimate intersection over union (IoU) between the input and ground truth masks regarding image segmentation.
While sharing a similar idea, it is non-trivial to extend and adapt their method to image registration.
For example, their method is limited to a static initial estimate, which requires a longer optimization trajectory than ours.
This may become a problem when applied to sport field registration, where the broadcast view change drastically even when there is small camera rotation.
We show through experiments that just having an error network is not enough, as we will show later in \tbl{errorobj}.

\section{Method}
\label{sec:method}

For clarity in presentation, we first assume that our models are pre-trained and detail our overall framework at inference time.
We then provide details on the training setup and the architectural choices.

\subsection{Inference}

\paragraph{Overview.}
Our pipeline is depicted in \fig{framework}. 
We assume a known planar sports field \template{} and undistorted images, so that we can represent the image-\template{} alignment with a homography matrix.
The framework can be broken down into two stages: the first stage provides an initial estimate of the homography matrix, the second iteratively optimizes this estimate. The first stage follows a typical feed-forward paradigm~\cite{Detone16,Tekin18a}, and we utilize a deep neural network.
However, any method can be used here instead, such as a database search~\cite{Sharma18d, Chen18e}.

The distinctiveness of our model comes from the second stage of the pipeline.
Using the first stage estimate, we warp the sports field \template{} to the current \view{}.
We concatenate this warped image with the current observed image, and evaluate the registration error through a second neural network.
We then backpropagate the estimated error to the 
the homography parameters to obtain the gradient, which gives the direction in which the parameters should be updated to minimize the registration error.
Then, using this gradient, we update the homography parameters.
This process is performed iteratively until convergence or until a maximum number of iterations is met.
This \emph{inference through optimization} allows our method to be significantly more accurate than a typical feed-forward setup, provided that our error model gives reasonable error predictions. 

\begin{figure}
\centering
\includegraphics[width=1.0\linewidth]{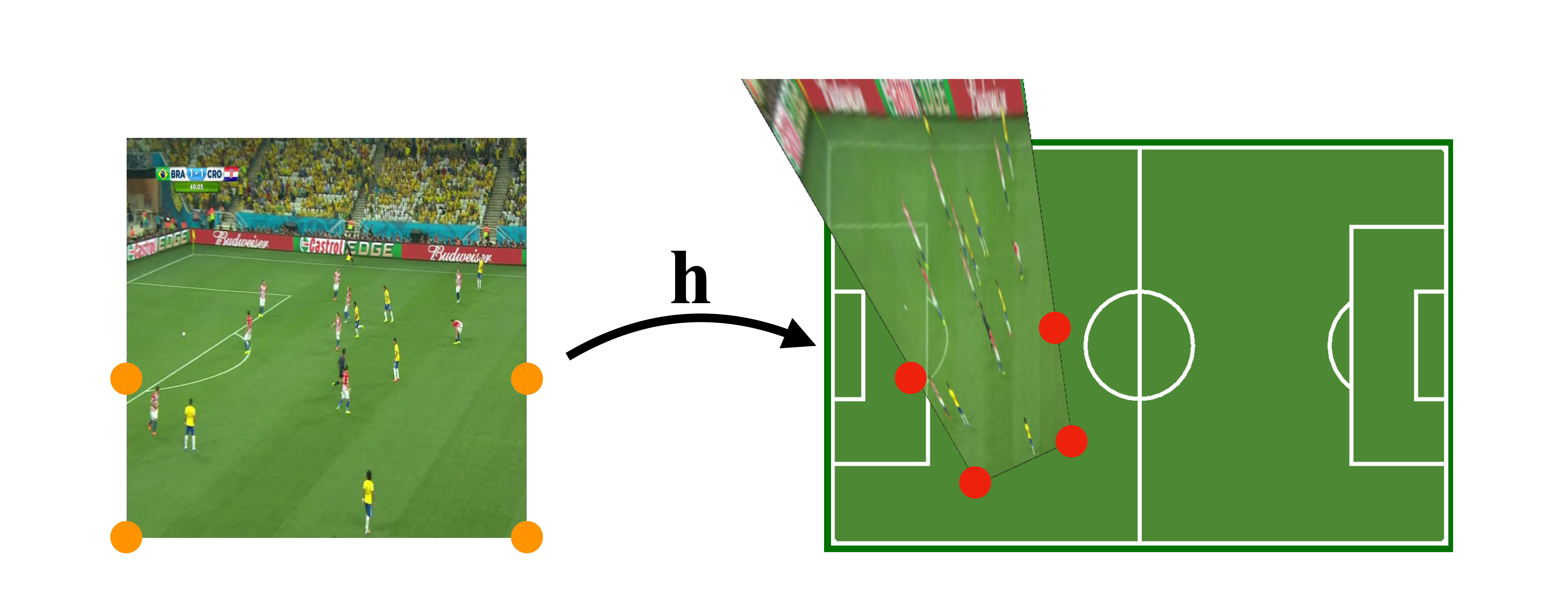}
\caption{
Illustration of control points.
The yellow dots on the left are the control points we use on the normalized image coordinate, and the red dots on the right are the control points after they are transformed via the homography $\bh$. 
Our initial registration network regresses the positions of the red dots.
}
\label{fig:controlpoints}
\end{figure}

\paragraph{Details -- initial registration.}
We follow the recent trend of using projected coordinates for pose parameterization~\cite{Detone16,Tekin18}. In the case of homographies, this can be done with 4 points~\cite{Baker06}.
We parameterize the homography $\bh$ defining the relationship between the input image $\bI$ and the target \template{} $\bm$ through the coordinate of the four control points on the current input image when warped onto the sports field \template{}.
Specifically, considering a normalized image coordinate system where the width and height of the image is set to one, and the centre of the image is at the origin, we use $(-0.5, 0.1)$, $(-0.5, 0.5)$, $(0.5, 0.5)$, and $(0.5, 0.1)$,  that is, the corners of the lower three-fifths of the image as our reference control points.
We write the reference control points $\bh_{ref}$ as,  
\begin{equation}
    \bh_{ref} = \left[-0.5, 0.1, -0.5, 0.5, 0.5, 0.5, 0.5, 0.1\right]^\top
    \;.
\end{equation}
We use the lower parts of the image as sports field broadcast videos are typically in a setup where the camera is looking down on the field, as shown in \fig{controlpoints}.

Let $(u_{k}, v_{k})$ denote the $k$-th control point of the current image $\bI$ projected onto the sports field \template{} $\bm$.
We then write the homography $\bh$ as
\begin{equation}
    \bh = \left[u_{1}, v_{1}, u_{2}, v_{2}, u_{3}, v_{3}, u_{4}, v_{4}\right]^\top
    \;.
\end{equation}
We obtain the actual transformation matrix $\bT$ from $\bh$ and $\bh_{ref}$ through direct linear transformation~\cite{Hartley00}.

Given an initial registration network $f_{\bPhi}\left(\cdot\right)$,
we obtain a rough homography estimate $\bhh^{(0)}$ for image $\bI$ as
\begin{equation}
    \bhh^{(0)} = f_{\bPhi}\left(\bI\right)
    \;,
\end{equation}
where the superscript in parenthesis denote the optimization iteration.

\paragraph{Details -- optimization.}
With the current homography estimate $\bhh^{(i)}$ at optimization iteration $i$, we warp the play-field \template{} $\bm$ to obtain an image of the \template{} in the current view, using a bilinear sampler~\cite{Jaderberg15} to preserve differentiability.
We concatenate the result of this warping operation 
$\cW\left(\bm, \bhh^{(i)}\right)$
and the image $\bI$, and pass it as input to the model
$g_{\bPsi}\left(\cdot\right)$ to obtain a prediction of the registration  error $\hat{\epsilon}^{(i)}$ as
\begin{equation}
    \hat{\epsilon}^{(i)} =
    g_{\bPsi}\left(\left[\bI; \cW(\bm, \bhh^{(i)})\right]\right)
    \;,
\end{equation}
where $\left[\;\;;\;\;\right]$ denotes concatenation along the channel direction of two images.
We then retrieve the gradient of $\hat{\epsilon}^{(i)}$ with respect to $\bhh^{(i)}$ and apply this gradient to retrieve an updated estimate.
In practice, 
we rely on Adam~\cite{Kingma15} for a stable optimization.

Note here that our registration error network is not trained to give updates.
It simply regresses to the correctness of the current estimate.
We show empirically in \refsec{ablation} that this is a much more effective than,
for example learning to provide a perfect homography, or learning to correct erroneous estimates.

\subsection{Training}
\label{sec:training}

To avoid overfitting,
we propose to purposely decouple the training of two networks. 
We show in \refsec{ablation} that this is necessary in order to obtain the best performance.

\paragraph{Initial registration network.}
To train the initial registration network, we directly regress the four control points of our \template{} warped into a given view using the ground truth homography.
With the ground truth homography $\bh_{gt}$, we train our deep network to minimize
\begin{equation}
    \loss_{init} = \left\|\bh_{gt} - \bhh^{(0)}\right\|_2^2
    = \left\|\bh_{gt} - f_\bPhi\left(\bI\right)\right\|_2^2
    \;.
\end{equation}
Note that while we use a deep network to obtain the initial homography estimate, \emph{any} other method can also be used in conjunction, such as nearest neighbor search.

\paragraph{Registration error network.}
To train the registration error network, we create random perturbations on the ground truth homography. 
We then warp the target \template{} to the view using the perturbed ground truth homography, and concatenate it with the input image to be used as input data for training. The network model is trained to predict a registration error metric, \eg the IoU. We detail our design choice of error metric in \refsec{ablation}.

In more detail, with the ground truth homography $\bh_{gt}$, we create a perturbed homography $\bh_{pert}$ by applying uniform noise hierarchically: one for global translation, and one for local translation of each control point.
Specifically, we add a global random translation $\alpha_{g} \sim U(-\delta_{g}, \delta_{g})$, where $\alpha_{g} \in \IR^2$, to all control points, and add a local random translation of $\alpha_{l} \sim U(-\delta_{l},\delta_{l})$, where $\alpha_{l} \in \IR^8$ 
individually to each control point.
We then warp the target \template{} according to the perturbed homography to create our input data for training.
Thus, the input to the registration error network for training is $\left[\bI; \cW\left(\bm, \bh_{pert}\right)\right]$.
Then, to train the network, we minimize
\begin{equation}
\begin{split}
    \loss_{error} =
    &
    \|
      \err\left(\bI, \cW(\bm, \bh_{pert})\right)\\
        &-
        g_\bPsi\left(\left[
            \bI ; \cW(\bm, \bh_{pert})
        \right]\right)
    \|_2^2
    \;,
\end{split}
\end{equation}
where $\err\left(\cdot,\cdot\right)$ is the error metric, for example the IoU value.

\section{Sports field registration results}
\label{sec:results}

We apply the proposed method to sports field registration.
We first discuss the datasets, baselines, the metrics used for our evaluation, as well as implementation details.
We then present qualitative and quantitative results of our method, compared to the state of the art.
We then provide experimental insights to our method.

\subsection{Experimental setup}

\begin{figure*}[h]
\includegraphics[width=1.0\linewidth]{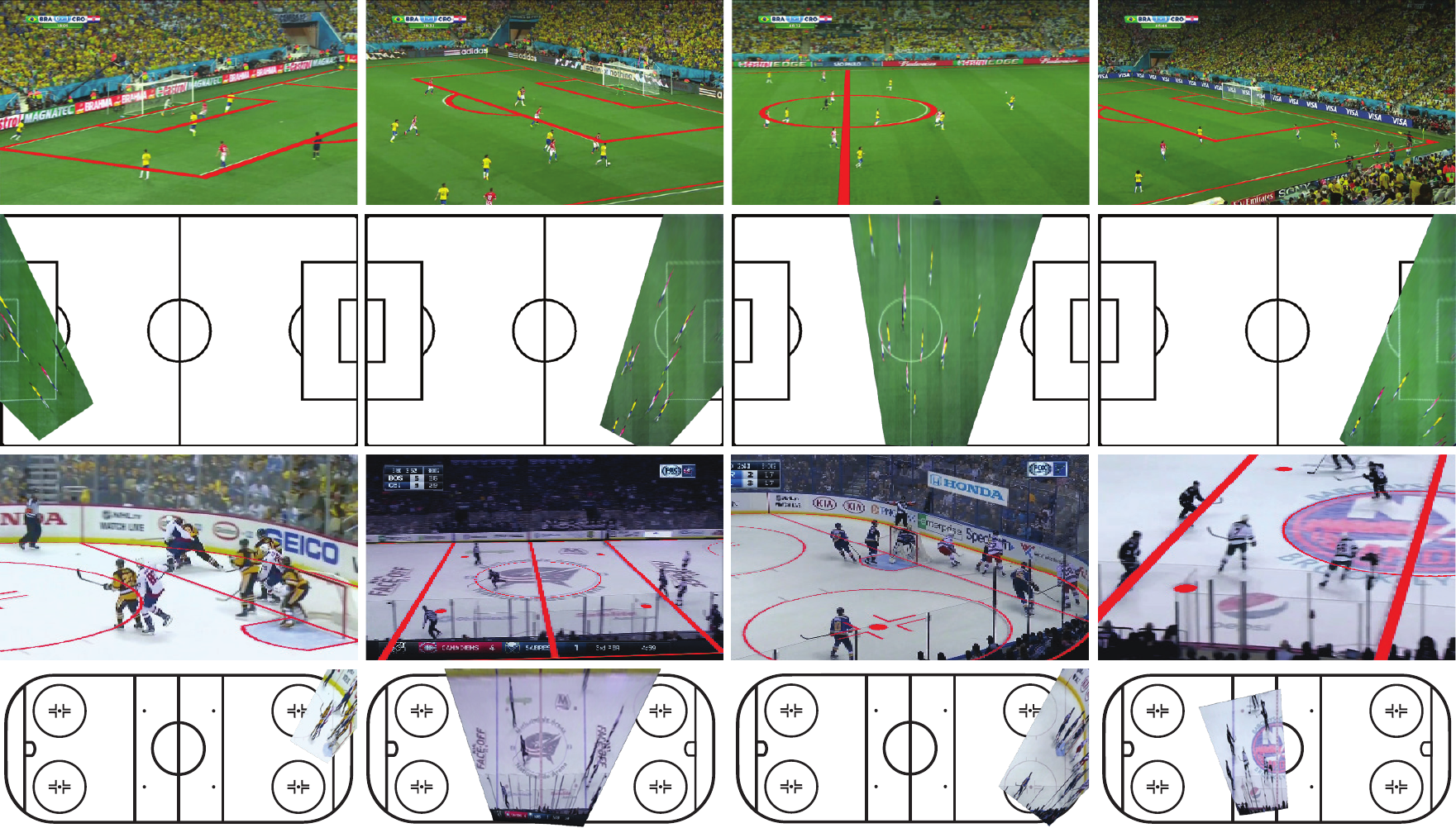}
\caption{
Qualitative highlights of our method. 
(Top) red lines are the sports field lines overlayed on the current view using estimated homographies. 
(Bottom) current view overlayed on sports field \template{}.
Our method can handle various sports fields and camera poses.
}
\label{fig:qualitative}
\end{figure*}

\paragraph{Datasets.}
To validate our method, we rely on two datasets.
The World Cup dataset~\cite{Homayounfar17} is a dataset made of broadcast videos of soccer games.
It has 209 images for training and validation, and 186 images for testing.
This dataset is extremely small, making it challenging to apply deep methods.
The state of the art for this dataset~\cite{Chen18e} relies on learning to transfer the input image to look similar to the sports field \template{}, then searching a database of known homographies and warped \templates{} to retrieve the estimate.
For our method, we use 39 images from the train-valid split as validation dataset, and respect the original test split for testing.
The Hockey dataset is 
composed of broadcast videos of NHL ice hockey games~\cite{Homayounfar17d}.
This is a larger dataset than the World Cup dataset, having 1.67M images in total.
Of this large dataset, we use two sequences of 800 consecutive images as validation and testing sets.
By using consecutive frames, we ensure that images from one game do not fall into different splits.
See \fig{qualitative} for example images.

\paragraph{Baselines.}
We compare our method against three existing works for sports field registration~\cite{Homayounfar17d, Sharma18d,Chen18e}.
As there is no publicly available implementation of the two methods~\cite{Homayounfar17d, Sharma18d}, we take the results reported on the respective papers for the World Cup dataset. 
For~\cite{Chen18e}, we use the authors' public implementation.
For \cite{Homayounfar17d} with the Hockey dataset, we use the reported results as a reference\footnote{
    No information is provided by the authors on how the the train, validation, and test splits are created, thus the results are not directly comparable. \label{ft:uot}
}.

In addition, we compare our method against feed forward baselines -- single stage feed-forward network ({\bf SSF}) and a two-stage feed-forward refinement network ({\bf FFR}).
We further explore whether the error registration network can be used alone by retrieving the initial estimate by searching a database of known poses, \eg the traing set, and using the example which gives the lowest error estimate.
We will refer to the initial estimate obtained through nearest neighbor search as {\bf NN}, and the fully optimized estimate as {\bf NNo}.
To do a nearest neighbor search we evaluate the registration error for the query image with all the training homographies using the trained registration error network, and return the homography with lowest estimated error.
Although this method is not scalable because the computational requirement grows linearly with the size of the database, it provides insight into the capability of the trained registration error network.

\paragraph{Metrics.}
As existing literature use different metrics~\cite{Homayounfar17d, Sharma18d,Chen18e},
IoU$_{part}$ and IoU$_{whole}$, we report both values.
IoU$_{part}$ is the intersection over union when only the visible region is considered, while IoU$_{whole}$ is the same considering the entire sports field \template{}.

\subsection{Implementation details}
\paragraph{Data augmentation.}
To avoid overfitting, we apply data augmentation.
(1) We crop the original image with random location and size. 
The ratio between the area of cropped image and the area of the original image is uniformly randomized to be between 90\% to 100\%.
(2) We apply random horizontal flips. 
(3) We create simulated shadows by overlaying a semi-transparent black patch with 50\% opacity.
To account for various settings, we apply random translation, rotation and scaling on the black patch. We uniformly randomly sample the translation between zero to half-width of the image, rotation between 0 to 45 degrees, and scaling factor between 0.5 to 2.
We further blur the edges with a Gaussian blur with kernel size 9 for smooth transition on borders.

\paragraph{Initial registration network.}
Following a recent trend~\cite{Ren15, He17},
we base our network on the ResNet-18 architecture~\cite{He16}.
Instead of the classification head, we simply replace the last fully connected layer to estimate 8 numbers which represent the homography, $\bh$. 
We use the pretrained weights for the network trained on ImageNet~\cite{Deng09}, and fine-tune.

\paragraph{Registration error network.}
For the registration error network, we also rely on the ResNet-18 architecture, but with spectral normalization~\cite{Miyato18} on all convolutional layers, and take as input a 6-channel image, that is, the concatenation of the input image and the warped target \template{}. Spectral normalization smooths the error predictions by constraining the Lipschitz constant of the model, which limits the magnitude of its gradients.
As the output of the registration error network cannot be negative, we use sigmoid function as the final activation function for the IoU-based error metrics, and squaring function for reprojection error metric.
For the registration network, as the input is very different from a typical image-based network, we train from scratch.

\paragraph{Hyperparameters.}
We train our networks with the Adam~\cite{Kingma15} optimizer, with default parameters $\beta_1=0.9$ and $\beta_2=0.999$, and with a learning rate of $0.0001$. 
We train until convergence,
and use the validation dataset to perform early stopping.
For the noise parameters $\delta_g$ and $\delta_l$ in \refsec{training} we empirically set $\delta_g = 0.05$ and $\delta_l = 0.02$, by observing the validation dataset results.
For inference, we again use Adam, but with a learning rate of $10^{-3}$.
We run our optimization for 400 iterations, and return the estimate that gave the lowest estimated error predicted by the trained registration error network.

\subsection{Results}
\label{sec:ablation}

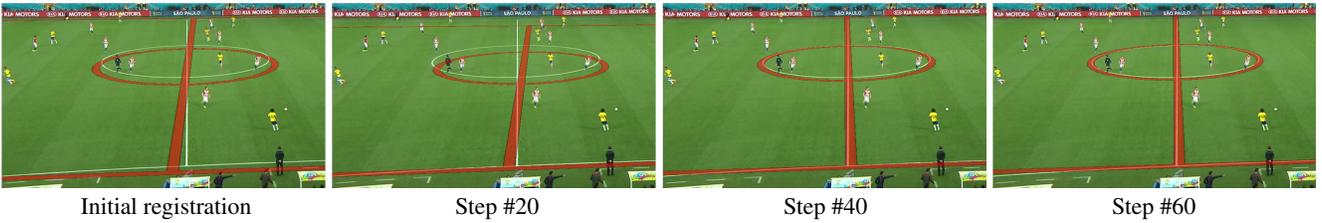
\begin{figure*}[h]
\centering
\begin{overpic} 
[width=\linewidth] 
{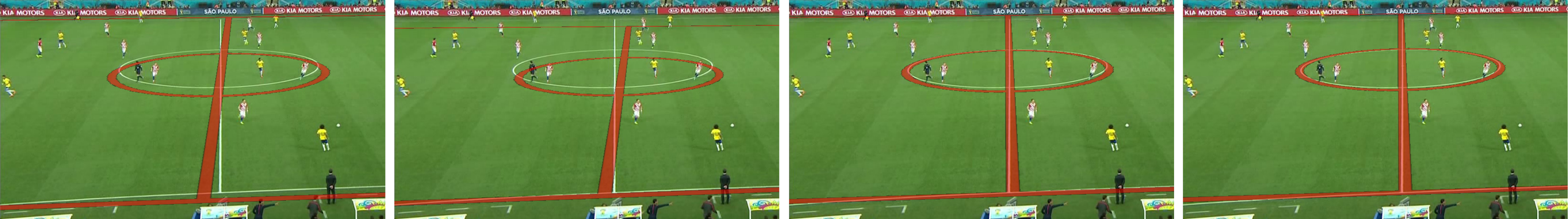}
\put(6,-2){\small{Initial registration}}
\put(34.5,-2){\small{Step \#20}}
\put(59.5,-2){\small{Step \#40}}
\put(84.5,-2){\small{Step \#60}}
\end{overpic}
\vspace{0.5em}
\caption{
Qualitative example demonstrating the effect of number of optimization iterations on registration accuracy.
From left to right, 
example registration result at iterations 0, 20, 40 and 60. 
Notice the misalignment near the center circle.
As more optimization iterations are performed, the registration becomes more accurate.
}
\label{fig:itervsiou}
\end{figure*}

\paragraph{Comparison against existing pipelines.}
Qualitative highlights are shown in \fig{qualitative} and \fig{itervsiou}, with quantitave results summarized in 
\tbl{worldcup}.
In \tbl{worldcup}, for the World Cup dataset, our method performs best in all evaluation metrics.
For the Hockey dataset, our method delivers near perfect results.

\paragraph{Comparison against feed-forward baselines.}
As shown in \tbl{worldcup}, having an additional feed-forward refinement network ({\bf FFR}) only provides minor improvement over the initial estimate ({\bf SFF}).
This phenomenon is more obvious in the WorldCup dataset results, where training data is scarce.
By contrast, our method is able to provide significant reduction in the registration error.

\newcolumntype{C}{>{\centering\arraybackslash}X}
\begin{table}
\begin{center}

\begin{tabularx}{\linewidth}{l l l C C C C C C}
    \toprule
      & & & {\cite{Homayounfar17d}}& {\cite{Sharma18d}} & {\cite{Chen18e}} &{\bf SFF} & {\bf FFR} & {Ours} \\
    \midrule
    \multirow{4}{*}{\rotatebox{90}{World Cup}} & \multirow{2}{*}{\rotatebox{90}{\scriptsize$\IoUwhole$}} & mean   & 83  & -- & 89.2 & 83.9 & 84.0 & {\bf89.8}\\
    & & median & -- & -- & 91.0  & 85.7 & 86.2 & \bf{92.9}\\
    \cmidrule[0.2pt]{2-9}
    & \multirow{2}{*}{\rotatebox{90}{\scriptsize$\IoUpart$}} & mean & -- & 91.4 & 94.7  & 90.2  & 90.3  & {\bf95.1} \\
    & & median & -- & 92.7 & 96.2  & 91.9  & 92.1  & {\bf96.7}\\
    \midrule
    \multirow{4}{*}{\rotatebox{90}{Hockey}} & \multirow{2}{*}{\rotatebox{90}{\scriptsize$\IoUwhole$}} & mean & 82\textsuperscript{\ref{ft:uot}}  & -- & -- & 86.5 & 93.0 & {\bf96.2} \\
    & & median & -- & -- & -- & 87.3 & 94.0 & {\bf97.0} \\
    \cmidrule[0.2pt]{2-9}
    & \multirow{2}{*}{\rotatebox{90}{\scriptsize$\IoUpart$}} & mean & -- & -- & -- & 90.4 & 96.0 & {\bf97.6}\\
    & & median & -- & -- & -- & 91.0 & 96.8 & {\bf98.4}\\
     \bottomrule
\end{tabularx}

\end{center}
\caption{
Quantitative results for different methods.
Best results are in bold. 
Our method performs best in all evaluation metrics.
See text for details.
}
\label{tbl:worldcup}
\end{table}

\def \figfive {0.23}

\begin{table}
\begin{center}

\begin{tabularx}{\linewidth}{lll C C C C C C}
    \toprule
      & & & {\scriptsize$\IoUwhole$}& {\scriptsize$\IoUpart$} & \scriptsize Reproj. & \scriptsize Coupled & {\scriptsize \bf NN} & {\scriptsize \bf NNo}\\
    \midrule
    \multirow{4}{*}{\rotatebox{90}{World Cup}} & \multirow{2}{*}{\rotatebox{90}{\scriptsize$\IoUwhole$}} & mean   & \bf 89.8  & 87.9 & 89.1 & 87.3 & 73.8 & 86.3\\
    & & median & \bf92.9 & 90.6 & 91.4 & 91.1 & 73.6 & 88.2\\
    \cmidrule[0.2pt]{2-9}
    & \multirow{2}{*}{\rotatebox{90}{\scriptsize$\IoUpart$}} & mean   & \bf95.1 & 94.7 & \bf95.1  & 94.4  & 87.4  & 94.0 \\
    & & median & \bf96.7 & 96.3 & 96.5  & 96.5  & 89.5  & 95.7\\
    \midrule
    \multirow{4}{*}{\rotatebox{90}{Hockey}} & \multirow{2}{*}{\rotatebox{90}{\scriptsize$\IoUwhole$}} & mean   & \bf96.2  & 95.6 & 94.9  & 87.9 & --    & -- \\
    & & median & \bf97.0 & 96.6 & 95.5  & 89.5 & -- & -- \\
    \cmidrule[0.2pt]{2-9}
    & \multirow{2}{*}{\rotatebox{90}{\scriptsize$\IoUpart$}} & mean   & \bf97.6 & 97.3 & 97.1  & 93.6  & -- & --\\
    & & median & \bf98.4 & 98.3 & 97.6  & 94.7  & -- & --\\
     \bottomrule
\end{tabularx}

\end{center}
\caption{
Quantitative results for different variants of our method.
Best results are in bold. 
$\IoUwhole$, $\IoUpart$, and Reproj. are three target error metrics we investigate. 
Coupled is when we couple the training of two networks.
{\bf NN} is when we use nearest neighbor search and {\bf NNo} is when we further optimize the homography estimate with the \errnet after {\bf NN}.
}
\label{tbl:errorobj}
\end{table}

\paragraph{Effect of different target error metrics.}
We also compare results when different target error is used for the training of the registration error network in \tbl{errorobj}.
We compare regressing to $\IoUwhole$, $\IoUpart$, and the average reprojection error of all pixels inside the current view (Reproj.).
Interestingly, regressing to $\IoUpart$ does not guarantee best performance in terms of $\IoUpart$. In all cases, regressing to $\IoUwhole$ gives best performance.

\paragraph{Coupled training.}
It is a common trend to train multiple components together.
However, our framework does not allow joint training, as the two networks are aiming for entirely different goals.
Nonetheless, we simultaneously trained the two networks, thus allowing the registration error network to see all the mistakes that the initial registration network makes during training (Coupled).
Coupled training, however, performs worse than decoupled training, as shown in \tbl{errorobj}.
In case of the Hockey dataset, coupled training performs even worse than feed-forward refinement.
This is because while the initial registration network is converging, it is making predictions with smaller and smaller mistakes, thus the registration error network is learning a narrow convergence basin due to the small perturbations it sees.  The estimates that fall out of the convergence basin can not be optimized using the learned error.
Therefore, it is necessary to have a decoupled training setup to stop this from happening.

\paragraph{Using only the error estimation network.}
The two variants, {\bf NN} and {\bf NNo}, provide insights into the capability of the registration error networks.
Due to the limited size of the database, i.e. training data, {\bf NN} provides initial estimates with lower accuracy than the single stage feed-forward network {\bf SFF}.
However, with optimization ({\bf NNo}), the registration results are even comparable to the results from our full pipeline.
This observation shows that the registration error network can provide a wide convergence basin, and can optimize for inaccurate initial estimates.

Note that we only test these methods on the World Cup dataset, as applying the method on Hockey dataset requires too much computation due to the larger database to search.

\paragraph{Inference performance.}
We perform all experiments on an Nvidia GTX 1080Ti GPU. To optimize one frame, our method achieves 41.76 optimization iterations per second, thus 9.58 seconds per frame. Our method also supports batch inference. To optimize a batch with 64 frames, it achieves 4.66 optimization iterations per second, thus in average 1.36 seconds per frame.

\subsection{Quality of the estimated error surface}

To validate that the trained registration error network can provide a convergence basin, we visualize the average estimated error surface for translation over all test samples.
To do so we create a regular grid with $X$ from $[-0.5, 0.5]$, and $Y$ from $[-0.5, 0.5]$ with resolution 50 by 50. 
For each point on the grid we warp the \template{} with ground truth homography combined with the translation from the origin to the point location. 
We then pass the observed image concatenated with the warped sports field to the trained registration error network, and infer the registration error at that point on the grid.
We calculate the error surface for all the test samples, and visualize the average.

\begin{figure}[h!]
\centering
\includegraphics[width=1.0\linewidth]{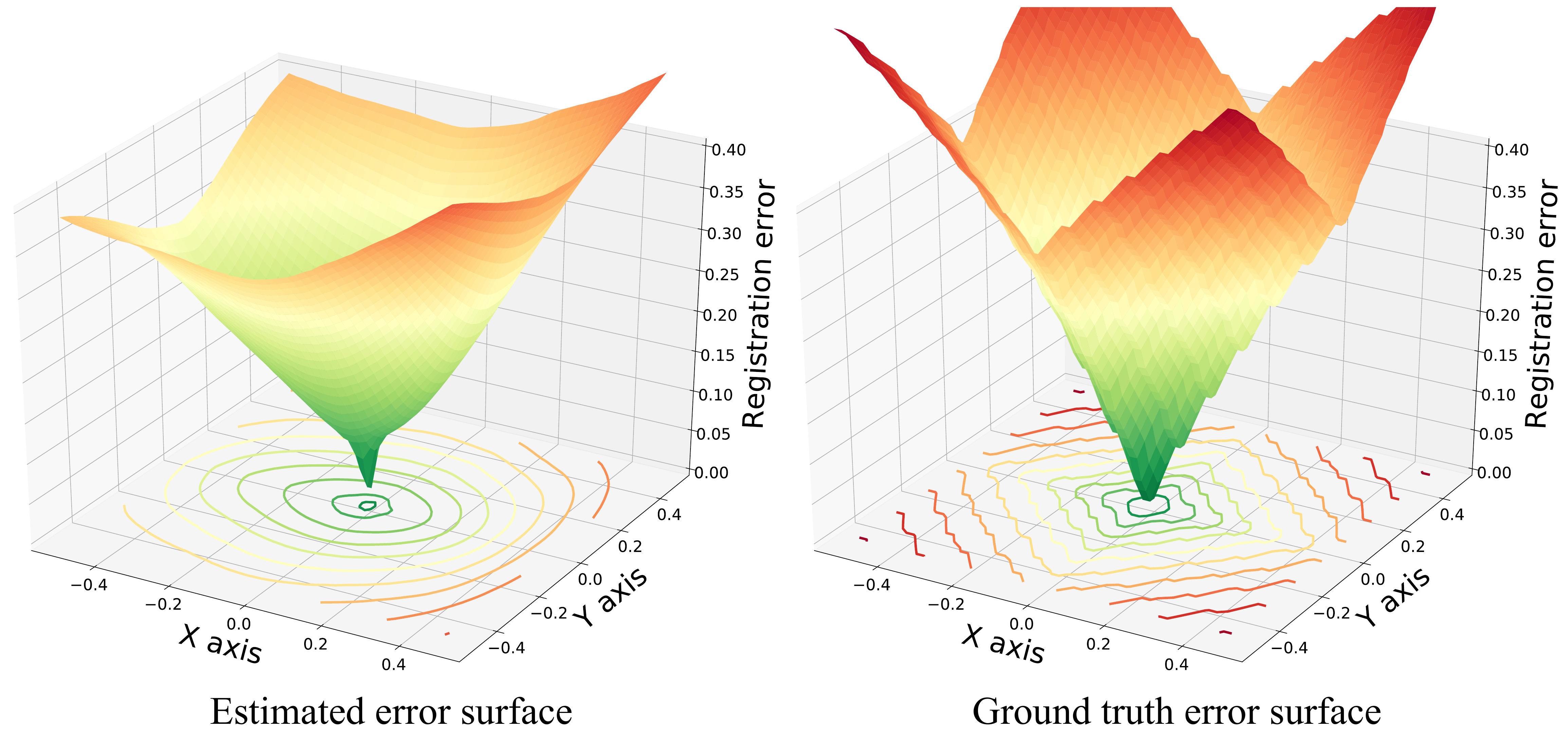}
\caption{
Average estimated and ground truth error surface visualization for translation.
See how the estimated error surface resembles the ground truth one, including the location of the minima at the centre.
This allows optimization through learned errors.
}
\label{fig:surface}
\end{figure}

As show in \fig{surface}, the estimated error surface resembles the ground truth one.
The error is lower towards the origin where the perturbation -- translation -- is smaller, and is higher towards the border where the perturbation is larger.
Most importantly, the minima of the estimated error is very close to the origin, which is the ground truth.
This allows our optimization-based inference to work properly.

\section{Toy experiment -- Line fitting}
\label{sec:motivation}

Beyond sport fields registration, our method could be applied to other tasks that involve parameter regression.
Here, we show briefly that, even a task as simple and generic as fitting a line equation in an image can benefit from our method.
We hope to shed some light into the potentials of our method.

Inspired by the experiment from DSAC~\cite{brachmann2017dsac, brachmann2018lessmore}, we validate our framework with the task of estimating the equation of a line from synthetic images, as shown in \fig{toy}.
Unlike DSAC, we are not learning to reject outliers via their pixel coordinates, but rather are directly regressing to the line equations given an image of a line.

\def \figtoy {1.0}
\begin{figure*}[h]
\includegraphics[width=\figtoy\linewidth]{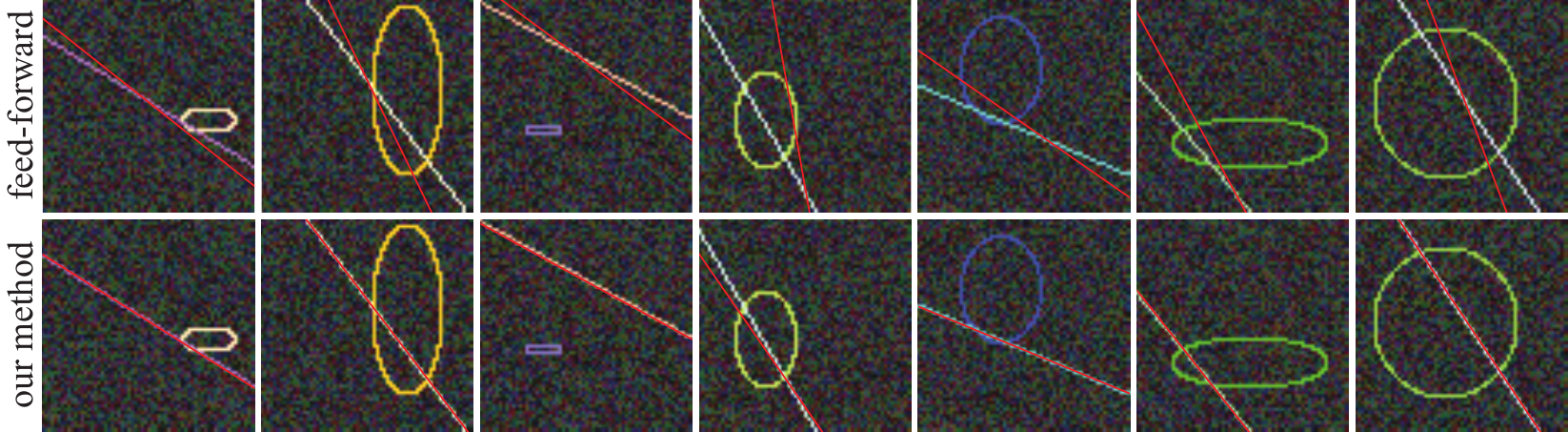}
\caption{
Estimating line equations of synthetic images.
The red line represents the estimated line equation from (top row) a feed-forward network, and (bottom row) the proposed method.
The other colored line in each image is the target line.
Our method provides accurate estimates, shown by the high overlap with the thick white line.
See \refsec{motivation} for details.
}
\label{fig:toy}
\end{figure*}

\subsection{Experimental setup} 
\paragraph{Initial network.}
We follow the same setup as our image registration task, but instead regress two parameters that define the equation of a line, that is, $a$ -- the angle and $b$ -- the intercept, where the line equation is given by $v = \tan(a)u + b$, and $u$ and $v$ are the image coordinates.

To create the synthetic images, we first generate random lines by selecting a random pivot point in an 64$\times$64 image, then uniformly sample in range $[-0.4\pi, 0.4\pi]$ to obtain its angle.
We draw this line with a random color.
We then add a random colored ellipse with random parameters as distraction, and finally apply additive Gaussian noise.
We use VGG-11~\cite{Simonyan15} as the backbone for the \regnet.

\paragraph{Error network.} We use the intercept error as the target error metric, that is maximum error between ground-truth and estimated intercept at $u=0$ or $u=63$.
To generate erroneous estimates for training, we add uniform noise $\alpha_{a}$ and $\alpha_{b}$ to the ground truth $a$ and $b$ respectively, where $\alpha_{a} \sim U[-0.1\pi, 0.1\pi]$, and $\alpha_{b} \sim U[-5, 5]$.
To render the estimate into an image in a differentiable way, we warp the the \template{} image which is simply an image of a line, using the hypothesized line parameters as in the case of image registration.
We concatenate the input image with the warp template to the error network to estimate the error, in this case the intercept error.
We also use VGG-11 as the backbone for the error network.

We train both networks until convergence and optimize for 400 iterations at inference time.

\subsection{Results}

As shown in \fig{toy}, our method estimates the line parameters more accurately than a feed-forward deep network.
Quantitative results are shown in \tbl{toy}.
As shown, even in this simple generic task, our method outperforms its feed-forward counterpart.
As this task can be viewed as a simplified version of other computer vision tasks, it shows that our method may be applicable outside the scope of the current paper.
We further highlight that this experimental setup is with unlimited labeled data.
Even in such a case, our method brings significant improvement in performance.

\newcolumntype{C}{>{\centering\arraybackslash}X}
\begin{table}[h!]
\begin{center}

\begin{tabularx}{\linewidth}{C C C}
    \toprule
    & Feed-forward & Ours \\
    \midrule
    mean error   & 5.1 & \bf{3.0}\\
    median error & 4.4 & \bf{1.5}\\
     \bottomrule
\end{tabularx}
\end{center}
\caption{
Quantitative results for line fitting.
Our method achieved better accuracy than a single stage feed-forward network.
This line fitting experiment can be viewed as a general regression task.
}
\label{tbl:toy}
\end{table}

\section{Conclusions}
\label{sec:conclusion}

We have proposed a two-stage pipeline for registering sports field \template{}s to broadcast videos accurately.
In contrast to existing methods that do single stage feed-forward inference, we opted for an optimization-based inference inspired by established classic approaches.
The proposed method makes use of two networks, one that provides an initial estimate for the registration homography, and one that estimates the error given the observed image and the current hypothesized homography.
By optimizing through the registration error network, accurate results were obtained.

We have shown through experiments that the proposed method can be trained with very sparse data, as little as 209 images, and achieve state-of-the-art performance.
We have further revealed how different design choices in our pipeline affect the final performance.
Finally, we have shown that our framework can be translated into other tasks and improve upon feed-forward strategies.

As future work, since the inference is optimization-based, we can naturally embed temporal consistency by reusing the optimization state for consecutive images to register sports field for a video.
We show preliminary results of doing so in our supplementary video.

\section*{Acknowledgements}

This work was partially supported by the Natural Sciences and Engineering
Research Council of Canada (NSERC) Discovery Grant ``Deep Visual Geometry
Machines'', NSERC Engage Grant ``Deep Localization and Modeling of Play-fields'', and by systems supplied by Compute Canada.

{\small
\bibliographystyle{ieee}
\bibliography{top}

\begin{thebibliography}{10}\itemsep=-1pt

\bibitem{Baker06}
S.~Baker, A.~Datta, and T.~Kanade.
\newblock {Parameterizing Homographies}.
\newblock Technical report, Robotics Institute, Carnegie Mellon University,
  2006.

\bibitem{Baker04b}
S.~Baker and I.~Matthews.
\newblock {Lucas-Kanade 20 Years On: A Unifying Framework}.
\newblock {\em International Journal of Computer Vision}, pages 221--255, 2004.

\bibitem{brachmann2017dsac}
E.~Brachmann, A.~Krull, S.~Nowozin, J.~Shotton, F.~Michel, S.~Gumhold, and
  C.~Rother.
\newblock Dsac-differentiable ransac for camera localization.
\newblock In {\em Proceedings of the IEEE Conference on Computer Vision and
  Pattern Recognition}, pages 6684--6692, 2017.

\bibitem{brachmann2018lessmore}
E.~Brachmann and C.~Rother.
\newblock Learning less is more-{6D} camera localization via {3D} surface
  regression.
\newblock In {\em CVPR}, 2018.

\bibitem{Tekin18}
T.~Bugra, S.~Sudipta~N, and F.~Pascal.
\newblock {Real-time Seamless Single Shot 6d Object Pose Prediction}.
\newblock In {\em Conference on Computer Vision and Pattern Recognition}, 2018.

\bibitem{Chen18e}
J.~Chen and J.~J. Little.
\newblock {Sports Camera Calibration via Synthetic Data}.
\newblock {\em Conference on Computer Vision and Pattern Recognition
  Workshops}, 2019.

\bibitem{Clark18}
R.~Clark, M.~Bloesch, J.~Czarnowski, S.~Leutenegger, and A.~J. Davison.
\newblock {LS-Net: Learning to Solve Nonlinear Least Squares for Monocular
  Stereo}.
\newblock {\em arXiv Preprint}, 2018.

\bibitem{Deng09}
J.~Deng, W.~Dong, R.~Socher, L.-J. Li, K.~Li, and L.~Fei-Fei.
\newblock {Imagenet: A Large-Scale Hierarchical Image Database}.
\newblock In {\em Conference on Computer Vision and Pattern Recognition}, 2009.

\bibitem{Detone16}
D.~DeTone, T.~Malisiewicz, and A.~Rabinovich.
\newblock {Deep image homography estimation}.
\newblock In {\em RSS Workshop on Limits and Potentials of Deep Learning in
  Robotics}, 2016.

\bibitem{DeTone17b}
D.~Detone, T.~Malisiewicz, and A.~Rabinovich.
\newblock {Superpoint: Self-Supervised Interest Point Detection and
  Description}.
\newblock {\em CVPR Workshop on Deep Learning for Visual SLAM}, 2018.

\bibitem{Ghanem12}
B.~Ghanem, T.~Zhang, and N.~Ahuja.
\newblock {Robust Video Registration Applied to Field-sports Video Analysis}.
\newblock In {\em International Conference on Acoustics, Speech, and Signal
  Processing}, 2012.

\bibitem{Gupta11}
A.~Gupta, J.~J. Little, and R.~Woodham.
\newblock {Using Line and Ellipse Features for Rectification of Broadcast
  Hockey Video}.
\newblock In {\em Canadian Conference on Computer and Robot Vision}, 2011.

\bibitem{Gygli17}
M.~Gygli, M.~Norouzi, and A.~Angelova.
\newblock Deep value networks learn to evaluate and iteratively refine
  structured outputs.
\newblock 2017.

\bibitem{Han18}
L.~Han, M.~Ji, L.~Fang, and M.~Nie{\ss}ner.
\newblock {RegNet: Learning the Optimization of Direct Image-to-Image Pose
  Registration}.
\newblock {\em arXiv Preprint}, 2018.

\bibitem{Hartley00}
R.~Hartley and A.~Zisserman.
\newblock {\em {Multiple View Geometry in Computer Vision}}.
\newblock Cambridge University Press, 2000.

\bibitem{He17}
K.~He, G.~Gkioxari, P.~Dollar, and R.~Girshick.
\newblock {Mask R-CNN}.
\newblock In {\em International Conference on Computer Vision}, 2017.

\bibitem{He15}
K.~He, X.~Zhang, R.~Ren, and J.~Sun.
\newblock {Delving Deep into Rectifiers: Surpassing Human-Level Performance on
  Imagenet Classification}.
\newblock In {\em International Conference on Computer Vision}, 2015.

\bibitem{He16}
K.~He, X.~Zhang, S.~Ren, and J.~Sun.
\newblock {Deep Residual Learning for Image Recognition}.
\newblock In {\em Conference on Computer Vision and Pattern Recognition}, 2016.

\bibitem{Homayounfar17d}
N.~Homayounfar, S.~Fidler, and R.~Urtasun.
\newblock {Sports Field Localization via Deep Structured Models}.
\newblock In {\em Conference on Computer Vision and Pattern Recognition}, 2017.

\bibitem{Homayounfar17}
N.~Homayounfar, S.~Fidler, and R.~Urtasun.
\newblock {Sports Field Localization via Deep Structured Models}.
\newblock In {\em Conference on Computer Vision and Pattern Recognition}, 2017.

\bibitem{Isola17}
P.~Isola, J.~Zhu, T.~Zhou, and A.~Efros.
\newblock {Image-To-Image Translation with Conditional Adversarial Networks}.
\newblock {\em Conference on Computer Vision and Pattern Recognition}, 2017.

\bibitem{Jaderberg15}
M.~Jaderberg, K.~Simonyan, A.~Zisserman, and K.~Kavukcuoglu.
\newblock {Spatial Transformer Networks}.
\newblock In {\em Advances in Neural Information Processing Systems}, 2015.

\bibitem{Jiang18b}
B.~Jiang, R.~Luo, J.~Mao, T.~Xiao, , and Y.~Jiang.
\newblock {Acquisition of Localization Confidence for Accurate Object
  Detection}.
\newblock In {\em European Conference on Computer Vision}, 2018.

\bibitem{Kehl17}
W.~Kehl, F.~Manhardt, F.~Tombari, S.~Ilic, and N.~Navab.
\newblock {SSD-6D: Making Rgb-Based 3D Detection and 6D Pose Estimation Great
  Again}.
\newblock In {\em International Conference on Computer Vision}, 2017.

\bibitem{Kendall15a}
A.~Kendall, M.~Grimes, and R.~Cipolla.
\newblock {Posenet: A Convolutional Network for Real-Time 6-DOF Camera
  Relocalization}.
\newblock In {\em International Conference on Computer Vision}, 2015.

\bibitem{Kingma15}
D.~Kingma and J.~Ba.
\newblock {Adam: {A} Method for Stochastic Optimisation}.
\newblock In {\em International Conference on Learning Representations}, 2015.

\bibitem{Lampert09}
C.~Lampert, M.~Blaschko, and T.~Hofmann.
\newblock {Efficient Subwindow Search: A Branch and Bound Framework for Object
  Localization}.
\newblock {\em IEEE Transactions on Pattern Analysis and Machine Intelligence},
  31:2129--2142, 2009.

\bibitem{Moreno09a}
V.~Lepetit, F.~Moreno-noguer, and P.~Fua.
\newblock {EPnP: An Accurate {O}(n) Solution to the PnP Problem}.
\newblock {\em International Journal of Computer Vision}, 81(2), 2009.

\bibitem{Lowe04}
D.~Lowe.
\newblock {Distinctive Image Features from Scale-Invariant Keypoints}.
\newblock {\em International Journal of Computer Vision}, 20(2):91--110, 2004.

\bibitem{Lucas81}
B.~Lucas and T.~Kanade.
\newblock {An Iterative Image Registration Technique with an Application to
  Stereo Vision}.
\newblock In {\em International Joint Conference on Artificial Intelligence},
  1981.

\bibitem{Marchand16}
E.~Marchand, H.~Uchiyama, and F.~Spindler.
\newblock {Pose Estimation for Augmented Reality: a Hands-on Survey}.
\newblock {\em IEEE Transactions on Visualization and Computer Graphics}, 2016.

\bibitem{Matas04}
J.~Matas, O.~Chum, M.~Urban, and T.~Pajdla.
\newblock {Robust Wide-Baseline Stereo from Maximally Stable Extremal Regions}.
\newblock {\em Image and Vision Computing}, 22(10):761--767, 2004.

\bibitem{Miyato18}
T.~Miyato, T.~Kataoka, M.~Koyama, and Y.~Yoshida.
\newblock {Spectral Normalization for Generative Adversarial Networks}.
\newblock In {\em International Conference on Learning Representations}, 2018.

\bibitem{Nguyen18d}
T.~Nguyen, S.~W. Chen, S.~S. Shivakumar, C.~J. Taylor, and V.~Kumar.
\newblock {Unsupervised Deep Homography: A Fast and Robust Homography
  Estimation Model}.
\newblock {\em IEEE Robotics and Automation Letters}, 2018.

\bibitem{Oberkampf96}
D.~Oberkampf, D.~DeMenthon, and L.~Davis.
\newblock {Iterative Pose Estimation Using Coplanar Feature Points}.
\newblock {\em Computer Vision, Graphics, and Image Processing},
  63(3):495--511, 1996.

\bibitem{Ono18}
Y.~Ono, E.~Trulls, P.~Fua, and K.~M. Yi.
\newblock {Lf-Net: Learning Local Features from Images}.
\newblock In {\em Advances in Neural Information Processing Systems}, 2018.

\bibitem{Puwein11}
J.~Puwein, R.~Ziegler, J.~Vogel, and M.~Pollefeys.
\newblock {Robust Multi-view Camera Calibration for Wide-baseline Camera
  Networks}.
\newblock In {\em IEEE Winter Conference on Applications of Computer Vision},
  2011.

\bibitem{Rad17}
M.~Rad and V.~Lepetit.
\newblock {Bb8: A Scalable, Accurate, Robust to Partial Occlusion Method for
  Predicting the 3D Poses of Challenging Objects Without Using Depth}.
\newblock In {\em International Conference on Computer Vision}, 2017.

\bibitem{Ren15}
S.~Ren, K.~He, R.~Girshick, and J.~Sun.
\newblock {Faster {R-CNN}: Towards Real-Time Object Detection with Region
  Proposal Networks}.
\newblock In {\em Advances in Neural Information Processing Systems}, 2015.

\bibitem{Ronneberger15}
O.~Ronneberger, P.~Fischer, and T.~Brox.
\newblock {{U-Net}: Convolutional Networks for Biomedical Image Segmentation}.
\newblock In {\em Conference on Medical Image Computing and Computer Assisted
  Intervention}, 2015.

\bibitem{Schonberger17}
J.~Sch\"{o}nberger, H.~Hardmeier, T.~Sattler, and M.~Pollefeys.
\newblock {Comparative Evaluation of Hand-Crafted and Learned Local Features}.
\newblock In {\em Conference on Computer Vision and Pattern Recognition}, 2017.

\bibitem{Sharma18d}
R.~A. Sharma, B.~Bhat, V.~Gandhi, and C.~V. Jawahar.
\newblock {Automated Top View Registration of Broadcast Football Videos}.
\newblock In {\em IEEE Winter Conference on Applications of Computer Vision},
  2018.

\bibitem{Mahendran17}
{Siddharth Mahendran and Haider Ali and Rene Vidal}.
\newblock 3d pose regression using convolutional neural networks.
\newblock In {\em The IEEE International Conference on Computer Vision (ICCV)
  Workshops}, 2017.

\bibitem{Simonyan15}
K.~Simonyan and A.~Zisserman.
\newblock {Very Deep Convolutional Networks for Large-Scale Image Recognition}.
\newblock In {\em International Conference on Learning Representations}, 2015.

\bibitem{Tang18}
C.~Tang and P.~Tan.
\newblock {Ba-Net: Dense Bundle Adjustment Network}.
\newblock In {\em International Conference on Learning Representations}, 2019.

\bibitem{Tekin18a}
B.~Tekin, S.~Sinha, and P.~Fua.
\newblock {Real-Time Seamless Single Shot 6D Object Pose Prediction}.
\newblock In {\em Conference on Computer Vision and Pattern Recognition}, 2018.

\bibitem{Verdie15}
Y.~Verdie, K.~M. Yi, P.~Fua, and V.~Lepetit.
\newblock {{TILDE}: A Temporally Invariant Learned {DEtector}}.
\newblock In {\em Conference on Computer Vision and Pattern Recognition}, 2015.

\bibitem{Wu07a}
F.~Wu and F.~Xiangyong.
\newblock {An Improved RANSAC Homography Algorithm for Feature Based Image
  Mosaic}.
\newblock In {\em Proceedings of the 7th WSEAS International Conference on
  Signal Processing, Computational Geometry \& Artificial Vision}, 2007.

\bibitem{Yan14}
Q.~Yan, Y.~Xu, X.~Yang, and T.~Nguyen.
\newblock {HEASK: Robust Homography Estimation Based on Appearance Similarity
  and Keypoint Correspondences}.
\newblock {\em Pattern Recognition}, 2014.

\bibitem{Yi16b}
K.~M. Yi, E.~Trulls, V.~Lepetit, and P.~Fua.
\newblock {LIFT: Learned Invariant Feature Transform}.
\newblock In {\em European Conference on Computer Vision}, 2016.

\bibitem{Xiang17}
X.~Yu, T.~Schmidt, V.~Narayanan, and F.~Dieter.
\newblock {PoseCNN: A Convolutional Neural Network for 6D Object Pose
  Estimation in Cluttered Scenes}.
\newblock {\em Robotics: Science and Systems (RSS)}, 2018.

\end{thebibliography}
}

\end{document}